\documentclass[runningheads]{llncs}

\usepackage[T1]{fontenc}
\usepackage{graphicx}
\usepackage{amsmath}
\usepackage{color}
\usepackage{multirow}
\usepackage{subcaption}
\usepackage{float}
\begin{document}

\title{CD-Net: Histopathology Representation Learning using Pyramidal Context-Detail Network}
%
\titlerunning{CD-Net: Pyramidal Context-Detail Network}
%

%
\author{Saarthak Kapse\inst{1}, Srijan Das\inst{2}, Prateek Prasanna\inst{1}}

\authorrunning{S. Kapse et al.}
%

\institute{Department of Biomedical Informatics, Stony Brook University, NY, USA
\and Department of Computer Science, Stony Brook University, NY, USA
\email{\{saarthak.kapse, prateek.prasanna\}@stonybrook.edu}}

\maketitle              

\vspace{-0.5cm}
\begin{abstract}

Extracting rich phenotype information, such as cell density and arrangement, from whole slide histology images (WSIs), requires analysis of large field of view, i.e more contexual information. This can be achieved through analyzing the digital slides at lower resolution. A potential drawback is missing out on details present at a higher resolution. To jointly leverage complementary information from multiple resolutions, we present a novel transformer based Pyramidal Context-Detail Network (CD-Net). CD-Net exploits the WSI pyramidal structure through co-training of proposed Context and Detail Modules, which operate on inputs from multiple resolutions. 
The residual connections between the modules enable the joint training paradigm while learning self-supervised representation for WSIs. The efficacy of CD-Net is demonstrated in classifying Lung Adenocarcinoma from Squamous cell carcinoma.

\let\thefootnote\relax\footnotetext{Code and trained models are made available at https://github.com/bmi-imaginelab/CD-Net}

\end{abstract}
\section{Introduction}
Advancements in computer vision have played a key role in histopathology whole slide image (WSI) analysis for crucial tasks like tumor grading, subtype classification, prognostic modeling, etc.
In order to process gigapixel WSIs using Deep Neural Networks (DNNs), they are divided into small patches. 
However, obtaining localized patch-level annotations of these WSIs is expensive and thus, only slide-level labels are generally available. To learn from such weak-supervision, Multiple Instance Learning (MIL) is adopted, in which the small patches are considered as a bag of instances (patches). A WSI is assigned to a positive bag if any of its patches is a positive instance. Finally, the patch level features are aggregated to predict the slide-level labels~\cite{DSMIL,feng2017deep,abmil}.
Note that traditional MIL models operate on the patch-level features extracted from a pre-trained DNN. Now the question remains, \textit{how do we learn high-fidelity discriminative representation of the WSI patches to facilitate robust functionality of the MIL models?}\\
In an attempt to emulate a pathologist's approach to analyzing WSIs, we note that patches at lower resolution provide more contextual information compared to the corresponding higher resolution patches (see Figure~\ref{fig:pipeline}a). However, these low resolution patches lack fine-grained details which are preserved in the corresponding patches at higher resolution.
Such fine-grained details, especially nuclei morphology, are crucial for tumor grading and classification. Note that `context' here refers to more global phenotype characteristics, such as distribution patterns of nuclei, co-occurrence patterns of cell types, etc. Thus, an approach towards utilizing both \textbf{contextual information} as well as the \textbf{fine-grained details} for analysing WSI is an intuitive research direction. This warrants the need of jointly training patches from multiple resolutions, which in turn scales up the memory requirement quadratically.
To this end, we present a novel architecture called \textbf{Context-Detail Network} (CD-Net) to learn holistic self-supervised representations exploiting the multiple resolutions of WSIs for disease classification. 

\noindent\textbf{Overview:} CD-Net is primarily composed of two modules -  a \textbf{context module} and a \textbf{detail  module} to jointly learn the complementary information from lower and higher resolutions of WSI patches. 
The context module operates on the patches at lower resolution whereas the detail module operates on the corresponding magnified patches from the higher resolution as shown in Figure~\ref{fig:pipeline}(b). 
In order to aggregate the fine-grained information from the magnified patches and combine them with the corresponding low resolution patch representations, the design choice of our architecture is based on vision transformer (ViT)~\cite{vit}. The detail module aggregates the fine-grained information within the image regions of the magnified patches through a self-attention mechanism. These aggregated features are then combined with the features of their corresponding low resolution patches. Finally, the context module operates on the low resolution patches to model the contextual information by computing interactions between any two image regions. Note that these image regions are sub-patches of a given patch of WSI which will be further detailed in Section~\ref{method}.
Thus, CD-Net leverages WSI pyramidal structure via joint training at multiple resolutions. This feature extractor is trained with self-supervision. For self-supervised learning (SSL), we adopt a transformer based learning framework, DINO~\cite{dino}, to learn discriminative representations for WSI patches. \\
To summarize, our major contributions are as follows: (1) we propose CD-Net that jointly learns contextual information and fine-grained details in WSIs by exploiting their multi-resolution pyramidal structure, and 
(2) our work is among the first to perform self-supervised representation learning of pure vision transformer models in histopathology. Our choice of DINO~\cite{dino} is motivated by the nature of features learned with DINO, i.e, features containing  semantic layout and, in particular, object boundaries for natural images. Similarly we believe that SSL features learned on WSIs would explicitly capture properties in whole slide patches, such as gland and cell distribution.
Notably, our approach has comparatively much lower computational overhead than processing larger patches at high resolution due to our design choice. We demonstrate the effectiveness of CD-Net on the publicly available TCGA-Lung cancer~\cite{tcga_luad,tcga_lusc,clark2013cancer} dataset in distinguishing Lung Adenocarcinoma (LUAD) from Squamous cell carcinoma (LUSC).


\section{Related work}
In this section, we briefly discuss self-supervised learning methods, multi-resolution networks, and vision transformers used in digital pathology tasks.

\noindent\textbf{Self-supervision in digital pathology.}
Contrastive learning methods in computational pathology have been explored in~\cite{ciga2022self,DSMIL,gildenblat2019self}, whereas non-contrastive learning methods for classical vision tasks have been explored in~\cite{boyd2021self,selfpath}. These self-supervised techniques have effectively alleviated the cost of large bags required in gigapixel MIL while demonstrating the robustness and generalizability of feature representations compared to pre-trained feature extractors from ImageNet. These techniques are suitable when there is a lack of localized annotation, and only slide-level labels are available. We use a non-contrastive self-supervision framework, DINO~\cite{dino}, as it is specifically designed for training ViTs.

\noindent\textbf{Multi-resolution networks in digital pathology.} Several studies such as ~\cite{msdamil,hooknet,DSMIL,bejnordi2015multi,tokunaga2019adaptive} have explored the rich multi-resolution information available in pyramidal WSIs. DSMIL~\cite{DSMIL} explored late fusion of features from multiple resolutions, whereas ~\cite{msdamil} utilized the patches from multiple resolutions in the same bag while performing MIL. However, these methods lack in learning joint representation from multiple resolutions. To address this, Rijthoven et al. recently proposed a hooking mechanism~\cite{hooknet} between two ``different'' encoder-decoder operating with input from lower and higher resolutions. This `hook' includes cropping of the contextual feature followed by concatenation with the feature maps in the target branch (high resolution branch), thereby aligning the pixels from multiple resolutions in a common semantic space. Such a mechanism is designed for the task of semantic segmentation and cannot be generalized for classification. In contrast, we aim at combining the contextual features with an aggregated detailed feature representation captured at a higher resolution. 

\noindent\textbf{Vision Transformer.}
Inspired from transformer-based self-attention models in machine translation~\cite{attention}, ViT~\cite{vit} is the first attempt of a convolution-free model in imaging datasets. Many variants of ViTs~\cite{deit,swin,crossvit,tnt} have been proposed, each of which surpasses SOTA CNNs in various vision tasks. Towards learning multi-scale representation of images, Transformers like TNT~\cite{tnt} and Cross-Vit~\cite{crossvit} have been proposed. In contrast, CD-Net exploits the pyramidal structure of the WSIs by operating on multi-resolution inputs.
Studies such as~\cite{gao2021instance,vu2022handcrafted,stegmuller2022scorenet} have explored variants of transformers for histopathology tasks. However, these studies do not explicitly leverage the rich multi-resolution information.

\section{Methodology}
\label{method}
In this section, we first describe the proposed CD-Net architecture which is illustrated in Fig~\ref{fig:pipeline}. Next, we describe the DINO framework which is used for self-supervised learning followed by MIL framework.


\noindent\textbf{Background.} For an in-depth understanding of the components in transformer such as MSA (Multi-head Self-Attention), LN (Layer Normalization), and MLP (Multi-Layer Perceptron), we refer the readers to ~\cite{attention}. In brief, the self-attention block explores the relationship between different regions in the image. MLP is applied between self-attention layers to transform the features and add non-linearity. LN~\cite{layernorm} is used to make the training stable and converge faster. 

\subsection{Context-Detail Network}

\begin{figure} [ht]
\begin{center}
\includegraphics[width=0.95\linewidth]{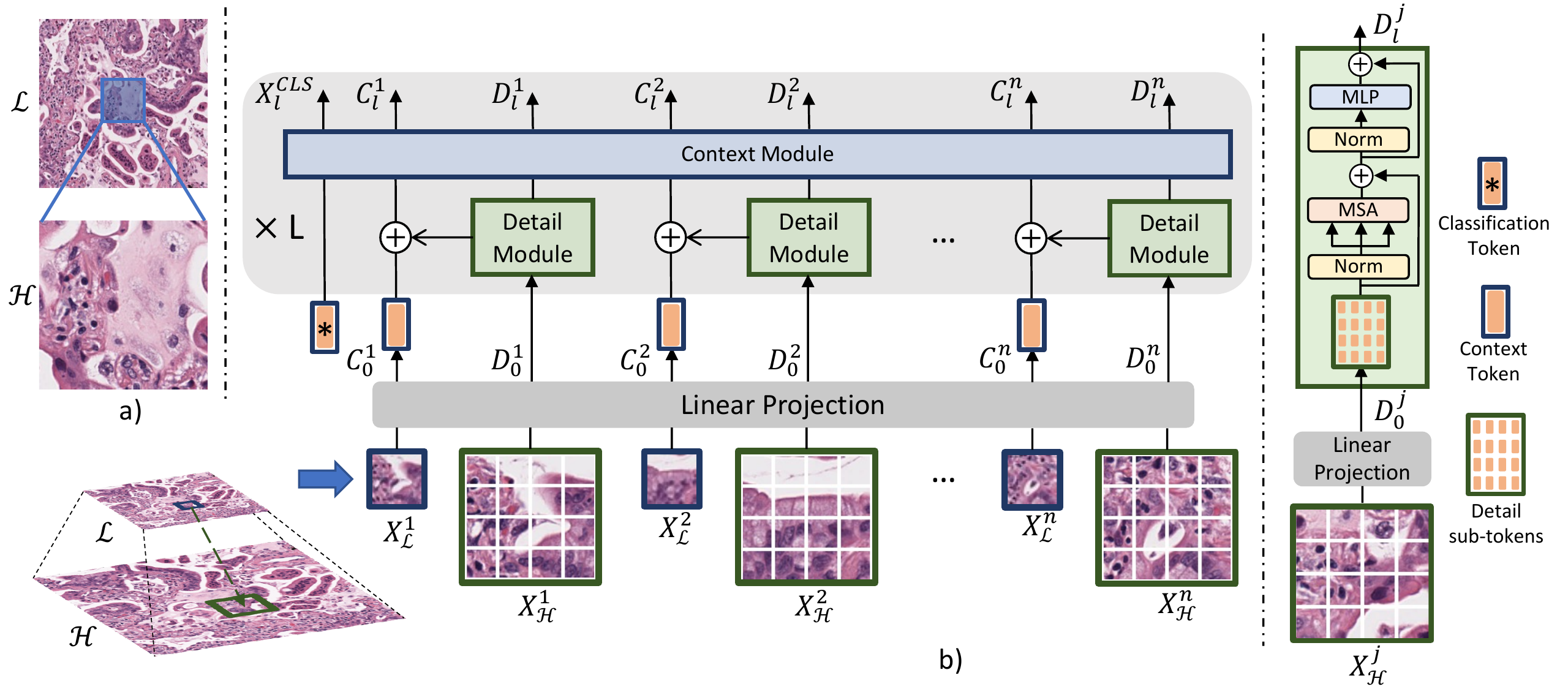}
\end{center}
   \caption[example] 
   { \label{fig:pipeline} a) An illustration of same dimensional patches from different resolutions $\mathcal{L}$ and $\mathcal{H}$. It may be observed that a patch at $\mathcal{L}$ has a larger field of view, whereas the one at $\mathcal{H}$ contains details about cell morphology. b) Illustration of the proposed \textbf{CD-Net}. 
    $C^j$, denoting the $j^{th}$ context token derived from the sub-patch at $\mathcal{L}$ (indicated by a \textcolor{blue}{blue} box), is input to the Context Module.
     $D^j$, denoting the $j^{th}$ detail sub-tokens derived from the corresponding magnified sub-patch at $\mathcal{H}$ (indicated by a \textcolor{green}{green} box), are input to the Detail Module. 
   }
\end{figure}

Each WSI $\mathcal{W}$ consists of a component image $\mathcal{W}_\mathcal{L}$ at lower resolution $\mathcal{L}$ and one image $\mathcal{W}_\mathcal{H}$ at higher resolution $\mathcal{H}$.
For each $\mathcal{W}_\mathcal{L}$, patches $w_1$, $w_2$, \dots $w_N$ are extracted by dividing the WSI, where $N$ is variable for each $\mathcal{W}_\mathcal{L}$. Each patch $w_i$, is decomposed into $n$ sub-patches $\mathcal{X}_\mathcal{L} = [X^1_\mathcal{L}, X^2_\mathcal{L}, ..., X^n_\mathcal{L}] \in \mathcal{R}^{n\times p\times p\times 3}$, which we call context patches, where ($p$, $p$) is the spatial size of each context patch. These context patches interact among each another to provide the implicit relationships among structured entities, such as cell clusters.
For context patches in $\mathcal{X}_\mathcal{L}$, we take the corresponding magnified patches $\mathcal{X}_\mathcal{H} = [X^1_\mathcal{H}, X^2_\mathcal{H}, ..., X^n_\mathcal{H}] \in \mathcal{R}^{n\times q\times q\times 3}$ from $\mathcal{H}$ as shown in Figure~\ref{fig:pipeline}b; we call them detail patches, where ($q$, $q$) is size of each detail patch. We note that the detail patch dimension $q$ is $\mathcal{H}/\mathcal{L}$ times the context patch dimension $p$. Each $\mathcal{X}_\mathcal{H}^j$ is further decomposed into $m$ sub-patches $\mathcal{X}^j_\mathcal{H} = [x^{j,1}_\mathcal{H}, x^{j,2}_\mathcal{H}, ..., x^{j,m}_\mathcal{H}] \in \mathcal{R}^{m\times s\times s\times 3}$, where ($s$, $s$) is the size of each sub-patch. 
The role of each $X^j_\mathcal{H}$ is to provide the high-magnification details to each $X^j_\mathcal{L}$. The sub-patches in each $X^j_\mathcal{H}$ are used to model the fine-grained information pertaining to $X^j_\mathcal{H}$. A linear projection then transforms each context patch $X^j_\mathcal{L}$ and each detail sub-patch $x^{j,k}_\mathcal{H}$ $\in$ $X^j_\mathcal{H}$, to context token ($C_0$) and detail sub-tokens ($D^j_0$) respectively. The tokenization of the context patches and the detail sub-patches follow 
\begin{equation} \label{patchembed}
C_0 = [X^1_\mathcal{L}\textbf{E}_\mathcal{\textbf{L}}; X^2_\mathcal{L}\textbf{E}_\mathcal{\textbf{L}};...,X^n_\mathcal{L}\textbf{E}_\mathcal{\textbf{L}}], \hspace{0.2in}
    D^j_0 = [x^{j,1}_\mathcal{H}\textbf{E}_\mathcal{\textbf{H}}; y^{j,2}_\mathcal{H}\textbf{E}_\mathcal{\textbf{H}};...,y^{j,m}_\mathcal{H}\textbf{E}_\mathcal{\textbf{H}}]
\end{equation}
where $\textbf{E}_\mathcal{\textbf{L}}$ and $\textbf{E}_\mathcal{\textbf{H}}$ are strided convolutional operations with $dim1$ and $dim2$ number of $p \times p$ and $s \times s$ filters respectively. 
The dimensions of the resultant context tokens and detail sub-tokens is ($n$, $dim1$) and ($m$, $dim2$), respectively. 
Subsequently, detail sub-tokens are fed into the detail module. This module consists of a transformer block which explores the interaction of detail sub-tokens as follows: 
\begin{equation}
   D'^{j}_l = D^j_{l-1} + MSA(LN(D^j_{l-1})); \hspace{0.2in} D^j_l = D^{'j}_{l} + MLP(LN(D^{'j}_{l}))
\end{equation}
where $l$ is index of the $l^{th}$ block of detail  module, and $j\in \{1,2,...n\}$.  Following the interaction step through attention mechanism (MSA), for each $D^j$, the corresponding $m$ detail sub-tokens are concatenated and passed through a linear projection, thus yielding an aggregated detail token. This token is then combined with the corresponding context token through a residual connection as
\begin{equation}
    C^j_{l-1} = C^j_{l-1} + FC(\mathrm{Concat}(D^j_l)), \hspace{0.1in} j\in \{1,2,...n\}
\end{equation}
This enables CD-Net to retain the information of both resolutions intact. 
Finally, the context module models the relationship between the context tokens, which is aided by the fine-grained information from the detail sub-tokens, using self-attention block:
\begin{equation}
  \mathcal{C}^{'}_l = \mathcal{C}_{l-1} + MSA(LN(\mathcal{C}_{l-1})); \hspace{0.2in} \mathcal{C}_l = \mathcal{C}^{'}_{l} + MLP(LN(\mathcal{C}^{'}_{l}))
\end{equation}
Thus in each Context-Detail block, the detail module models the interaction between the magnified sub-token and passes on the fine-grained details to corresponding context tokens. Consequently, these context tokens interact with other context tokens in the context module to learn representations for each patch $w_i$. CD-Net is composed of $L$ stacked blocks of both detail module and context module. Also, a classification token $X^{CLS}$ is appended with $C_0$ in Equation~\ref{patchembed}, which serves as the image representation.

\noindent\textbf{Position encoding.} To retain spatial information of image, learnable 1D position encoding is assigned while tokenizing context patches and detail sub-patches in CD-Net (see Equation~\ref{patchembed}). Note that separate position encoding is assigned for context tokens $PC \in \mathcal{R}^{{(n+1)}\times dim1}$ and detail sub-tokens $PD \in \mathcal{R}^{m\times dim2}$. For brevity we have abstained from mentioning the position encoding in equations.

\vspace{-4mm}

\subsection{Self-supervision of CD-Net}
The non-contrastive DINO~\cite{dino} framework consists of a student and a teacher branch. Both the networks share the same architecture with different sets of parameters. The student is trained to match the output of a given teacher network. Whereas, the teacher is trained using an exponential moving average (EMA) on the student weights. Thus the teacher branch may be closer to the role of the mean teacher used in self-training~\cite{meanteacher}. 

For training this framework in a batch, for each patch $w_i$, two random set of augmentations are applied, producing $w_i^1$ and $w_i^2$. One is fed to student branch, and the other to teacher branch. 
Given a fixed teacher network, the framework learns to match the probability distributions of the student and the teacher by minimizing cross-entropy loss w.r.t. the parameters of student network. In this framework both the student and teacher adopt the CD-Net architecture for learning multi-resolution representation for each patch of WSI.

\subsection{Multiple Instance Learning (MIL)}
Multiple instance learning is a well-established and widely used method in WSI analysis, we refer the readers to ~\cite{DSMIL,chikontwe2020multiple,feng2017deep,abmil} for an overview. In this work, we used the DSMIL~\cite{DSMIL} framework. Followed by the self-supervision, the trained models are used to extract the features for each patch, which are then provided to the MIL model for learning slide-level classification.

\section{Experimental Design}
We now present our experimental results on a clinical WSI dataset.
For this study, we fix $\mathcal{L}$ and $\mathcal{H}$ to magnifications 5$\times$ and 20$\times$ respectively. We set the size of each patch $w_i$ from $\mathcal{W}_\mathcal{L}$ to $224 \times 224$. The dataset is divided into separate train and test set as mentioned in next subsection. Note that WSIs from the entire train set are only used for self-supervision learning. This train set is further divided into train and validation sets for MIL model training. Thus, we ensure that the test set remains independent from both self-supervised feature extractor as well as MIL model, to ensure fair evaluation.

\noindent\textbf{Dataset Description.}
TCGA Lung Cancer dataset comprises of 1042 diagnostic digital slides~\cite{tcga_luad}. This dataset contains two sub-types of lung cancer, LUAD and LUSC, and the task is to classify these two sub-types. We split the WSIs in 828 training (LUAD : 425, LUSC : 403) and 214 testing (LUAD : 105, LUSC : 109) samples consistent with~\cite{DSMIL}. There is a slight discrepancy in dataset split as few slides are no longer available. Following patch extraction, the $\mathcal{L}$ resolution yields 0.9 million patches and $\mathcal{H}$ yields 12 million patches. 

\noindent\textbf{Baselines.} 
As a baseline, we first evaluated the performance of single-resolution ViTs on the histopathology dataset. We trained the ViT model using DINO separately on two resolutions, $\mathcal{L}$ and $\mathcal{H}$. 
We also trained TNT~\cite{tnt} on patches from $\mathcal{W}_\mathcal{L}$ that models local information within an image patch through interaction within its sub-patches in the same resolution. 
This experiment shows the need of modeling fine-grained information from higher resolution patches (say $\mathcal{H}$), and thus the importance of joint training of multiple resolution WSI patches. 
For the multi-resolution learning, we established a baseline through the late fusion of feature vectors from above single-resolution trained models followed by construction of feature pyramid, inspired  by~\cite{DSMIL}. 
It should be noted that the MIL model is trained separately for features from single resolution, late fused multi-resolution features, and for our proposed jointly learned multi-resolution features.


\noindent\textbf{CD-Net Parameters.} For baseline, we adapt ViT-S~\cite{deit} and TNT-S~\cite{tnt} architecture on TCGA dataset. 
Details about CD-Net network architecture and its training parameters can be found in the supplementary.
For self-supervised training, we use the default learning rate and other parameters from DINO~\cite{dino}.


\noindent\textbf{MIL Parameters.} For training the DSMIL~\cite{DSMIL} framework, we use a learning rate of 0.0002 with a batch size of 1 to handle variable bag size. We train the MIL model for 40 epochs and use a weight decay of 0.05 to prevent over-fitting. All other MIL parameters are same as in~\cite{DSMIL}.


\section{Results}
\begin{figure} [ht]
\begin{center}
\includegraphics[width=.9\linewidth]{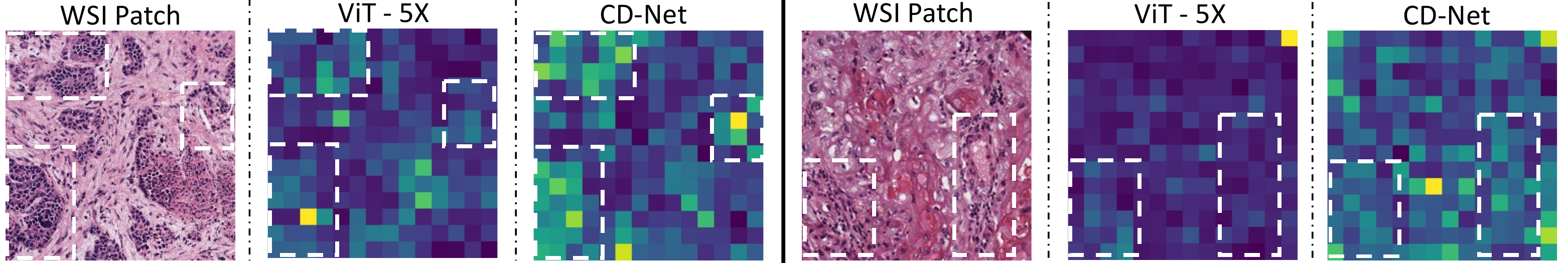}
\end{center}
   \caption[example] 
   { \label{fig:vis_1}
    Attention maps of single resolution ViT vs. multi-resolution CD-Net.
    The attention map from  CD-Net appear denser pertaining to fine grained details it gets from $\mathcal{H}$. White bounding box represents the area with high cell density.
   }
\end{figure}

Table~\ref{tab:results1} shows the classification performance of CD-Net, single-resolution ViTs and their late fusion. We also compare our proposed CD-Net with TNT~\cite{tnt} trained at 5$\times$. 
In Table~\ref{tab:results1}, we observe that the ViT models trained with patches from $\mathcal{L}$ are more effective than the models trained with patches from $\mathcal{H}$. This substantiates the need of larger field of view for this classification task.
Towards multiple-resolution models, the late-fusion of 5$\times$ and 20$\times$ features performs similar to the models trained with single-resolution in this dataset. In contrast, our CD-Net outperforms all the representative baselines as it takes into account the similar field of view as ViT at $\mathcal{L}$, and the fine-grained details from $\mathcal{H}$ which enhances the representation learning capabilities of the model. We also provide a qualitative visualization of the transformer attention maps of CD-Net and ViT for the same WSI patch in Figure~\ref{fig:vis_1}. The maps seems to suggest that CD-Net is more attentive to local details such as regions with high cellular density compared to ViT at $\mathcal{L}$.

\noindent \textbf{Complexity Analysis.} To be noted, single resolution ViT models trained on 5$\times$ and 20$\times$ both have input patches of 224$\times$224. Whereas, CD-Net effectively takes the patch from 224$\times$224 from 5$\times$ and the corresponding magnified 896$\times$896 patch from 20$\times$ as shown in Figure~\ref{fig:pipeline}. On the contrary, one may argue the use of 896$\times$ 896 patches to train a single resolution transformer yielding features similar to that of CD-Net. But this would increase the computation by 100 times compared to our CD-Net.

\noindent\textbf{Ablation Study.} First, we present the importance of Self-Supervised Learning (SSL) by comparing the results of feature extractor ResNet-18~\cite{resnet} pretrained on ImageNet~\cite{imagenet} (ResNet IN) against SSL on respective patches in Table~\ref{tab:results2}. 
Also, we compare our results with a study close to our work - DSMIL~\cite{DSMIL}, which used SimCLR~\cite{simclr} to train ResNet-18 architecture (ResNet SSL)  on the TCGA Lung Cancer dataset. We extract features per WSI patch from their provided model weights for both resolutions to feed to MIL. The superior performance of ViT SSL shows the discriminative power of our vision transformer based architecture and its SSL strategy (DINO) compared to the previous baselines.

\begin{table}[ht]
    \caption{a) Comparison of joint multiple-resolution training with single-resolution training. LF denotes Late Fusion.  b) Ablation Study for showing the effectiveness of transformer-based SSL on TCGA dataset.}
    \begin{subtable}{.5\linewidth}
      \centering
\begin{tabular}{|c|c|c|c|} 
\hline
Resolution                                                 & Model  & Accuracy $\uparrow$ & AUC $\uparrow$   \\ 
\hline
\multirow{2}{*}{5$\times$~}                                  & ViT    & 0.902    & 0.953  \\ 
\cline{2-4}
                                                      & TNT    & 0.897    & 0.958  \\ 
\hline
20$\times$~                                                  & ViT    & 0.864    & 0.943  \\ 
\hline
\begin{tabular}[c]{@{}c@{}}LF \\~5$\times-20\times$~\end{tabular} & ViT    & 0.902    & 0.954  \\ 
\hline
\begin{tabular}[c]{@{}c@{}}Joint\\5$\times-20\times$\end{tabular} & CD-Net & \textbf{0.911}    & \textbf{0.958}  \\
\hline
\end{tabular}
\caption{\label{tab:results1}}
    \end{subtable}%
    \scalebox{0.9}{
    \begin{subtable}{.5\linewidth}
      \centering
\begin{tabular}{|c|c|c|c|} 
\hline
Resolution                                                                & Model             & Accuracy $\uparrow$ & AUC $\uparrow$    \\ 
\hline
\multirow{3}{*}{5$\times$~}                                                 & ResNet IN & 0.715    & 0.749  \\ 
\cline{2-4}
                                                                     & Resnet SSL       & 0.831    & 0.911  \\ 
\cline{2-4}
                                                                     & ViT SSL~              & 0.902    & 0.953  \\ 
\hline
\multirow{3}{*}{20$\times$~}                                                & ResNet IN  & 0.724    & 0.760  \\ 
\cline{2-4}
                                                                     & Resnet SSL       & 0.775    & 0.884  \\ 
\cline{2-4}
                                                                     & ViT SSL~              & 0.864    & 0.943  \\ 
\hline
\multirow{2}{*}{\begin{tabular}[c]{@{}c@{}}LF\\$5\times-20\times$~\end{tabular}} & Resnet SSL       & 0.789    & 0.891  \\ 
\cline{2-4}
                                                                     & ViT SSL~              & 0.902    & 0.954  \\
\hline

\end{tabular}
\caption{\label{tab:results2}}
    \end{subtable}}
\end{table}

Interestingly, CD-Net when evaluated on Camelyon16~\cite{camelyon16}, outperforms ViT at $\mathcal{L}$ but performs sub-par compared to ViT at $\mathcal{H}$. This is because, for metastasis detection in~\cite{camelyon16}, details such as nuclei morphology are potentially more informative rather than context; ViT at $\mathcal{H}$ performs better than ViT at $\mathcal{L}$. Here, the sub-par performance of CD-Net is owing to the lower embedding dimension $dim2$ for detail sub-tokens than the embedding dimension $dim1$ for context tokens. Thus, CD-Net under-represents the detail sub-patches through detail module. The results can be found in supplementary. This presents an interesting opportunity to selectively invoke multi-resoluton/single-resolution models based on prior clinical knowledge.

\section{Conclusion}
In this work, we presented CD-Net, a novel architecture to exploit the pyramidal structure of WSIs through co-training of context and detail module operating at lower and higher resolutions, respectively. CD-Net can effectively represent the larger field of view with context module, and concurrently the fine-grained details through detail module. We demonstrate the effectiveness of CD-Net in tumor-subtype classification on TCGA Lung cancer dataset. We believe CD-Net is more suitable for diagnostic tasks where relatively lower resolution pathology features such as glands, cell distribution patterns, etc. are vital, and the contribution of higher resolution cellular-scale features is less crucial but still relevant. Our future work will focus on developing a co-training multi-resolution model which can overcome current limitations and can generalize to other diagnostic tasks. 
%
%
%
%





\bibliography{main} 

\begin{thebibliography}{10}
\providecommand{\url}[1]{\texttt{#1}}
\providecommand{\urlprefix}{URL }
\providecommand{\doi}[1]{https://doi.org/#1}

\bibitem{tcga_luad}
Albertina, B., et~al.: Radiology data from the cancer genome atlas lung
  adenocarcinoma [tcga-luad] collection. The Cancer Imaging Archive  (2016)

\bibitem{layernorm}
Ba, J.L., Kiros, J.R., Hinton, G.E.: Layer normalization. arXiv preprint
  arXiv:1607.06450  (2016)

\bibitem{bejnordi2015multi}
Bejnordi, B.E., et~al.: A multi-scale superpixel classification approach to the
  detection of regions of interest in whole slide histopathology images. In:
  Medical Imaging 2015: Digital Pathology. SPIE (2015)

\bibitem{camelyon16}
Bejnordi, B.E., et~al.: Diagnostic assessment of deep learning algorithms for
  detection of lymph node metastases in women with breast cancer. Jama  (2017)

\bibitem{boyd2021self}
Boyd, J., et~al.: Self-supervised representation learning using visual field
  expansion on digital pathology. In: Proceedings of the IEEE/CVF International
  Conference on Computer Vision (2021)

\bibitem{dino}
Caron, M., et~al.: Emerging properties in self-supervised vision transformers.
  In: Proceedings of the IEEE/CVF International Conference on Computer Vision
  (2021)

\bibitem{crossvit}
Chen, C.F.R., et~al.: Crossvit: Cross-attention multi-scale vision transformer
  for image classification. In: Proceedings of the IEEE/CVF International
  Conference on Computer Vision (2021)

\bibitem{simclr}
Chen, T., et~al.: A simple framework for contrastive learning of visual
  representations. In: International conference on machine learning. PMLR
  (2020)

\bibitem{chikontwe2020multiple}
Chikontwe, P., et~al.: Multiple instance learning with center embeddings for
  histopathology classification. In: International Conference on Medical Image
  Computing and Computer-Assisted Intervention. Springer (2020)

\bibitem{ciga2022self}
Ciga, O., et~al.: Self supervised contrastive learning for digital
  histopathology. Machine Learning with Applications  (2022)

\bibitem{clark2013cancer}
Clark, K., et~al.: The cancer imaging archive (tcia): maintaining and operating
  a public information repository. Journal of digital imaging  (2013)

\bibitem{vit}
Dosovitskiy, A., et~al.: An image is worth 16x16 words: Transformers for image
  recognition at scale. arXiv preprint arXiv:2010.11929  (2020)

\bibitem{feng2017deep}
Feng, J., et~al.: Deep miml network. In: Proceedings of the AAAI Conference on
  Artificial Intelligence (2017)

\bibitem{gao2021instance}
Gao, Z., et~al.: Instance-based vision transformer for subtyping of papillary
  renal cell carcinoma in histopathological image. In: International Conference
  on Medical Image Computing and Computer-Assisted Intervention. Springer
  (2021)

\bibitem{gildenblat2019self}
Gildenblat, J., et~al.: Self-supervised similarity learning for digital
  pathology. arXiv preprint arXiv:1905.08139  (2019)

\bibitem{tnt}
Han, K., et~al.: Transformer in transformer. Advances in Neural Information
  Processing Systems  (2021)

\bibitem{msdamil}
Hashimoto, N., et~al.: Multi-scale domain-adversarial multiple-instance cnn for
  cancer subtype classification with unannotated histopathological images. In:
  Proceedings of the IEEE/CVF conference on computer vision and pattern
  recognition (2020)

\bibitem{resnet}
He, K., et~al.: Deep residual learning for image recognition. In: Proceedings
  of the IEEE conference on computer vision and pattern recognition (2016)

\bibitem{abmil}
Ilse, M., et~al.: Attention-based deep multiple instance learning. In:
  International conference on machine learning. PMLR (2018)

\bibitem{tcga_lusc}
Kirk, S., et~al.: Radiology data from the cancer genome atlas lung squamous
  cell carcinoma [tcga-lusc] collection. The Cancer Imaging Archive  (2016)

\bibitem{selfpath}
Koohbanani, N.A., et~al.: Self-path: Self-supervision for classification of
  pathology images with limited annotations. IEEE Transactions on Medical
  Imaging  (2021)

\bibitem{DSMIL}
Li, B., et~al.: Dual-stream multiple instance learning network for whole slide
  image classification with self-supervised contrastive learning. In:
  Proceedings of the IEEE/CVF Conference on Computer Vision and Pattern
  Recognition (2021)

\bibitem{swin}
Liu, Z., et~al.: Swin transformer: Hierarchical vision transformer using
  shifted windows. In: Proceedings of the IEEE/CVF International Conference on
  Computer Vision (2021)

\bibitem{imagenet}
Russakovsky, O., et~al.: Imagenet large scale visual recognition challenge.
  International journal of computer vision  (2015)

\bibitem{stegmuller2022scorenet}
Stegm{\"u}ller, T., et~al.: Scorenet: Learning non-uniform attention and
  augmentation for transformer-based histopathological image classification.
  arXiv preprint arXiv:2202.07570  (2022)

\bibitem{meanteacher}
Tarvainen, A., et~al.: Mean teachers are better role models: Weight-averaged
  consistency targets improve semi-supervised deep learning results. Advances
  in neural information processing systems  (2017)

\bibitem{tokunaga2019adaptive}
Tokunaga, H., et~al.: Adaptive weighting multi-field-of-view cnn for semantic
  segmentation in pathology. In: Proceedings of the IEEE/CVF Conference on
  Computer Vision and Pattern Recognition (2019)

\bibitem{deit}
Touvron, H., et~al.: Training data-efficient image transformers \& distillation
  through attention. In: International Conference on Machine Learning. PMLR
  (2021)

\bibitem{hooknet}
Van~Rijthoven, M., et~al.: Hooknet: Multi-resolution convolutional neural
  networks for semantic segmentation in histopathology whole-slide images.
  Medical Image Analysis  (2021)

\bibitem{attention}
Vaswani, A., et~al.: Attention is all you need. Advances in neural information
  processing systems  (2017)

\bibitem{vu2022handcrafted}
Vu, Q.D., et~al.: Handcrafted histological transformer (h2t): Unsupervised
  representation of whole slide images. arXiv preprint arXiv:2202.07001  (2022)

\end{thebibliography}
\bibliographystyle{splncs04} 

%

\title{Supplementary Material}
\author{Saarthak Kapse$^1$, Srijan Das$^2$, Prateek Prasanna$^1$}
\authorrunning{S. Kapse et al.}
%

\institute{Department of Biomedical Informatics, Stony Brook University, NY, USA
\and Department of Computer Science, Stony Brook University, NY, USA
\email{\{saarthak.kapse, prateek.prasanna\}@stonybrook.edu}}

\maketitle              
\vspace{-0.6cm}
\begin{table}[H]
    \centering
    \caption{\textbf{Camelyon16 results.} Comparison of joint multiple-resolution training with single-resolution training. LF denotes Late Fusion. ViT at 20$\times$ performed significantly better compared to ViT at 5$\times$. This may be attributed to the fact that, for metastasis detection in Camelyon16, details such as nuclei morphology (more apparent on 20$\times)$ are potentially more informative compared to the context. The sub-par performance of CD-Net is possibly due to the much lower embedding dimension $dim2$ for detail sub-tokens than the embedding dimension $dim1$ for context tokens. Thus, CD-Net under-represents the detail sub-patches through detail module, thereby hampering the performance in this dataset.}
      
\begin{tabular}{|c|c|c|c|} 
\hline
\textbf{Resolution}   & \textbf{Model}  & \textbf{Accuracy} $\uparrow$ & \textbf{AUC} $\uparrow$   \\ 
\hline
5$\times$~          & ViT    & 0.742    & 0.723  \\ 
\hline
20$\times$~         & ViT    & 0.890    & 0.940  \\ 
\hline
~ LF 5$\times-20\times$~ & ViT    & 0.860    & 0.939  \\ 
\hline
Joint 5$\times-20\times$ & CD-Net & 0.757    & 0.792  \\
\hline
\end{tabular}

\end{table}

\vspace{-1.2cm}
\begin{table}[H]
\centering
    \caption{\textbf{Description of CD-Net parameters.} head1 is number of Multi-head in self-attention block assigned in context module, and head2 is number of Multi-head in self-attention block assigned in detail module. MLP r is the mlp ratio between each transformer block. Other parameters are described in the paper.}

\begin{tabular}{|c|c|c|c|c|c|c|c|c|c|c|c|} 
\hline
\textbf{Dataset}           & \textbf{L} & \textbf{dim1} & \textbf{head1} & \textbf{dim2} & \textbf{head2} & \textbf{p}  & \textbf{q}  & \textbf{s}  & \textbf{n}   & \textbf{m}  & \textbf{MLP r}  \\ 
\hline
TCGA Lung Cancer~ & 12    & 384  & 6     & 24   & 4     & 16 & 64 & 16 & 196 & 16 & 4      \\ 
\hline
Camelyon16        & 12    & 512  & 8     & 24   & 4     & 16 & 64 & 16 & 196 & 16 & 4      \\
\hline
\end{tabular}

\end{table}

\vspace{-0.9cm}

\begin{table}[H]
\centering
    \caption{Comparison of number of tokens and self-attention (SA) computation needed in each SA block for i) ViT on image of size 224$\times$224, ii) ViT on image with size 896$\times$896, iii) CD-Net effectively taking both 224$\times$224 from 5$\times$ and 896$\times$896 from 20$\times$. Image is decomposed into patches of size 16$\times$16. As may be observed, with local SA in CD-Net in the detail module, the effective computation is reduced by more than 100 times compared to (ii)}

\begin{tabular}{|c|c|c|c|} 
\hline
                           & \textbf{ViT 224$\times$224}    & \textbf{ViT 896$\times$896 }           & \textbf{CD-Net (224$\times$224, 896$\times$896)}         \\ 
\hline
Number of tokens           & 196              & 196$\times$16                  & 196 + 196$\times$16                      \\ 
\hline
Self-Attention & 196$^2$=38K & (196$\times$16)$^2$=9.8M & 196$^2$ + 196$\times$(16$^2$)=88K  \\
\hline
\end{tabular}

\end{table}

\vspace{-4cm}
\begin{figure}
\begin{center}
\includegraphics[width=.9\linewidth]{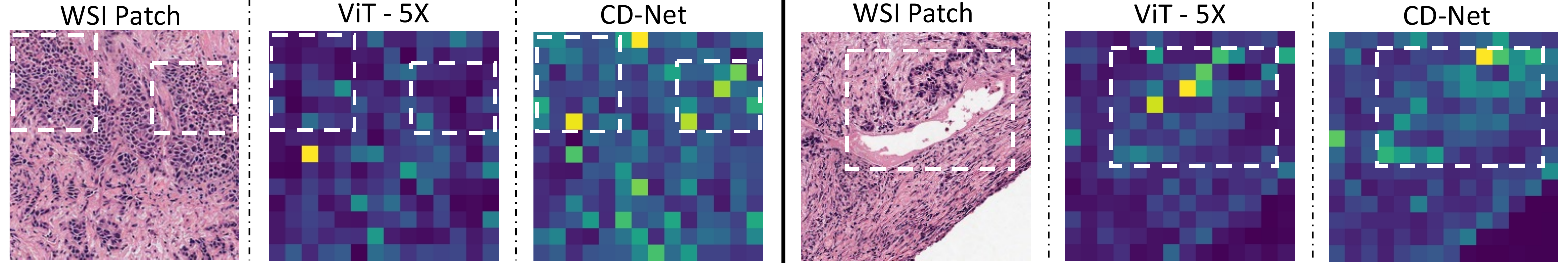}
\end{center}
   \caption[example] 
   { \label{fig:vis_2}
    More illustrations of Attention maps of single resolution ViT vs. multi-resolution CD-Net.
    The attention map from  CD-Net appear denser pertaining to fine grained details it gets from $\mathcal{H}$. White bounding box represents the area with high cell density.
   }
\end{figure}

\end{document}